\documentclass[10pt,twocolumn,letterpaper]{article}

\usepackage[pagenumbers]{cvpr} %
\usepackage{tabularx}
\usepackage{multirow}%
\usepackage{makecell}%
\usepackage{subcaption}
\graphicspath{{figs/}}

\definecolor{cvprblue}{rgb}{0.21,0.49,0.74}
\usepackage[pagebackref,breaklinks,colorlinks,allcolors=cvprblue]{hyperref}

\title{The GENEA Challenge 2026: A Large-Scale Disentangled Evaluation of Speech-Driven Gesture Generation on the Seamless Interaction Dataset}
\author{
Rajmund Nagy\textsuperscript{1},
Silvia Arellano García\textsuperscript{1},
Hendric Voss\textsuperscript{2},
Mihail Tsakov\textsuperscript{3},
\\
Taras Kucherenko\textsuperscript{4},
Youngwoo Yoon\textsuperscript{5},
Gustav Eje Henter\textsuperscript{1,6}
\\[2ex]
\textsuperscript{1}KTH Royal Institute of Technology
\textsuperscript{2}Bielefeld University
\\
\textsuperscript{3}Independent researcher
\textsuperscript{4}National Library of Sweden
\\ 
\textsuperscript{5}Electronics and Telecommunications Research Institute (ETRI)
\textsuperscript{6}Motorica AB
}

\begin{document}
\maketitle

\begin{abstract}
    This preprint presents the results of the fourth GENEA Challenge, in which we performed large-scale human evaluation on five speech-driven gesture-generation systems trained by participating teams on the Seamless Interaction dataset, containing speech and motion data for dyadic conversations. Similarly to the 2023 GENEA Challenge, we used a disentangled evaluation methodology to assess motion quality and alignment with the speech without confounding effects between the two, and also performed a dyadic mismatching study to isolate the effect of listening and reacting to the interlocutor. We additionally introduce a new semantic gesture-generation task using the Grounded Gestures subset of the data, and propose a new semantic evaluation methodology using text mismatching. In total, we ran four large-scale user studies, collecting over $23,000$ votes from $869$ test-takers.

    The user study on motion realism compared silent motion clips from different sources (a participating system or motion capture). The dataset's filtered segments had substantially higher motion quality than all challenge submissions (68--95\% pairwise winrate); the top two submissions were closely matched ($54\%$ vs $46\%$ head-to-head winrate); and the remaining three submissions showed a clear hierarchy.

    The user study measuring the alignment between motion and input speech used the audio mismatching methodology of the GENEA Leaderboard. Here, the motion-capture segments provided a conceptual ceiling for the results at 62\% speech-motion alignment score, with the top-performing submission significantly behind at 32\%, and the remaining four submissions only slightly above the $0\%$ expected bottom-line performance of an input-independent system.

    In the third study, we evaluated the ability of the systems to generate motion with dyadic adaptations to the interlocutor. Here, the motion-capture segments (which were filtered to contain dyadic adaptations) again provided a ceiling for the results at $65\%$ appropriateness score, but none of the challenge submissions scored substantially above chance level, indicating that the systems were not yet able to generate motion that is responsive to the interlocutor.

    Finally, the proposed semantic mismatching evaluation found that the dataset contained highly expressive gestures in the Grounded Gestures subset, as test-takers could identify the matching sentence transcript 79\% of the time. However, once again, almost all challenge submissions were unable to generate semantically expressive motion, with the best-performing submission only achieving an $8\%$ appropriateness score.

    The collected votes and outputs will be made publicly available on the challenge website (\href{https://genea-workshop.github.io/2026/challenge/}{https://genea-workshop.github.io/2026/challenge/}) to facilitate reproducibility and further research.
\end{abstract}

\section{Introduction}
\label{sec:intro}
Automatic gesture generation is an active research area that still lacks standardised evaluation practices, making it difficult to compare competing systems. The GENEA Challenges \cite{kucherenko2021large, kucherenko2024evaluating,kucherenko2023challenge} have been running for several years, with the goal of improving evaluation practices and determining the level of progress in automatic gesture generation. Most recently, the GENEA Leaderboard \cite{nagy2026towards} initially evaluated six published gesture-generation systems on the widely used BEAT2 dataset, and found that the gap between the motion realism of the motion-capture data and the best-performing systems was closing, but the systems were still unable to generate motion that was consistently aligned with the speech. Later, the leaderboard was updated to include a seventh system -- a diffusion transformer from the authors of Seamless Interaction \cite{agrawal2025seamless} -- which was able to achieve parity with motion-capture data in both realism and speech-gesture alignment. Therefore, the leaderboard established a new state of the art in gesture generation, and the BEAT2 dataset has become insufficient to measure further progress in the field.

With the GENEA Challenge 2026, we propose to instead use the recently released Seamless Interaction dataset \cite{agrawal2025seamless} to track progress, due to its much larger size, and its focus on dyadic interactions and semantically expressive gestures. The size of the dataset and the more complex nature of the interactions mean that adapting previous gesture-generation approaches to this dataset may not be straightforward. Additionally, the higher density of motion-capture artefacts in Seamless Interaction compared to BEAT2 may also make preprocessing and filtering more challenging than before. However, as shown by the aforementioned state-of-the-art diffusion transformer submission to the GENEA Leaderboard, by training on the Seamless dataset, it is possible to achieve groundbreaking results (as the model reached mocap-level alignment in a mismatching evaluation for the first time) even in a cross-dataset evaluation.

Hosting the challenge in conjunction with the first Interactive Social Avatars (ISA) workshop at ECCV, we invited submissions trained on the Seamless Interaction dataset. In total, we collected five submissions from four teams, which we then evaluated against each other and against the motion-capture data. This preprint presents the evaluation setup and the results of the subjective evaluations; system-description papers containing more details on the challenge submissions will be released on the workshop website at \href{https://interactive-social-avatars.github.io/}{https://interactive-social-avatars.github.io/}.

\subsection{Terminology}
\label{sec:terminology}
Throughout this preprint, we use the following terms:
\begin{itemize}
    \item A \emph{condition} is a particular motion source, i.e., either a challenge submission or the motion-capture data.
    \item The authors of the challenge submissions are referred to as \emph{challenge participants}.
    \item The evaluators recruited for our user studies are referred to as \emph{test-takers}.
\end{itemize}

\subsection{Challenge submissions}

The challenge evaluated five models: GestFlow \cite{guillen2026gestflow}; a flow-matching model from the UNICAMP team \cite{gomez2026unified}; a submission from the UM-FERI team \cite{crnek2026umferi}; and the DyaSync model alongside a secondary semantic variant (DyaSync-sem) \cite{fang2026dyasync}. For more details on the challenge submissions, please refer to the individual system description papers.

\section{Dataset}
\label{sec:dataset}

The challenge uses the recently released Seamless Interaction dataset \cite{agrawal2025seamless}. In particular, we leverage the 3400-hour \emph{dyadic conversations} subset, containing recordings of over 4000 actors having either natural or prompt-guided dialogue, and the 380-hour \emph{grounded gestures} subset, where actors take turns expressing prompted phrases with their gestures and improvising a natural response.

For the evaluation, challenge participants submitted generated motion for 473 short segments selected from a combination of public and private test splits of the dataset \cite{agrawal2025seamless}. In the case of nondeterministic challenge submissions, we collected five different outputs for each segment, and we selected a random sample for each segment, to prevent cherry-picking and detect mode collapse.

\section{Methodology}
\label{sec:methodology}
Following the spirit of previous GENEA Challenges \cite{kucherenko2021large, kucherenko2024evaluating,kucherenko2023challenge} and the GENEA Leaderboard \cite{nagy2026towards}, we use a disentangled evaluation methodology to assess different aspects of gesture generation in isolation from each other. Concretely, we define the following four evaluation tasks:

\begin{enumerate}
    \item[T1] \textbf{Motion realism}: How natural and visually convincing is the generated motion?
    \item[T2] \textbf{Alignment with speech}: Is the generated motion aligned with the input speech?
    \item[T3] \textbf{Alignment with the interlocutor}: Is the generated motion responsive to the interlocutor?
    \item[T4]  \textbf{Semantic alignment}: Does the generated motion express the meaning of a highlighted keyword?
\end{enumerate}

The first two evaluation tasks may be considered the \emph{core} evaluation of gesture generation, while the last two assess more advanced (and lesser studied) capabilities of gesture-generation systems. T1, T2, and T3 use the \textit{dyadic conversations} subset, while T4 uses the \textit{grounded gestures} subset of the data. In T1, T2, and T4, the model receives an audio clip and its transcript, and generates the speaker's motion. In T3, it additionally receives the interlocutor's audio, transcript, and motion, and generates the agent's motion.

\subsection{Task-dependent methodology}
\label{sec:task-methodology}
Each user study uses a pairwise voting format to keep task complexity manageable and to simplify the numerical evaluation, but stimulus pairs and response options are constructed differently. Below, we present a brief overview of each evaluation setup; for more details, refer to \cite{nagy2026towards} (regarding T1 and T2) and \cite{kucherenko2023challenge} (regarding T3).

\subsubsection{Construction of stimulus pairs}
In T1, we evaluate the visual quality of motion without considering alignment with the speech. Correspondingly, on every user-study screen we compare two silent videos rendered from the outputs of different conditions for the same input speech.

In T2, we measure alignment between speech and motion without considering the visual quality of the gestures. To this end, we use the audio mismatching evaluation \cite{nagy2026towards}, where user-study screens show two videos of the same motion but with two different audio segments -- one matched, and one mismatched (i.e., taken from another segment). This design removes the confounding effect of motion quality, in contrast to the naive setup of comparing motion clips from different systems. Test-takers are instructed to vote on whether the motion fits the rhythm and timing of the speech, emphasises the appropriate parts of the utterance, captures the semantic content of the speech, or reflects its emotional expression more accurately.

In T3, we use the dyadic mismatching methodology \cite{kucherenko2023challenge} to evaluate whether the motion of the agent appears attentive and responds appropriately to the interlocutor in their conversation, without considering visual quality or alignment to the speech. Mismatched segments were collected from challenge participants by providing them with additional mismatched test-set inputs -- in particular, for every matched dyadic test-set segment, we provided a mismatched counterpart where the interlocutor's motion and speech were taken from another, unrelated segment. In this user study, the only difference within a matched-mismatched video pair is the agent's motion; the interlocutor's motion and both speakers' audio are kept the same. To facilitate the evaluation, the agent is always displayed on the left side of the screen and the interlocutor on the right. Similarly, the agent's speech is played through the left audio channel and the interlocutor's speech through the right channel.

In T4, we use short sentences from the \emph{grounded gestures} subset to evaluate the semantic expressivity of the motion. We propose a novel text mismatching methodology, similar to the audio mismatching in T2, in which test-takers are shown a single muted, monadic video together with two candidate sentences -- one matched (the sentence spoken in the video) and one mismatched (drawn from a random segment). Test-takers are asked to select the sentence they believe the speaker is gesturing for, or indicate that the gestures are suitable for both or neither sentence.

\subsection{Segment selection}
\label{sec:segment_selection}
Our user studies present short rendered videos, ranging from 9--20 seconds in T1 and T2, 16--20 seconds in T3, and 2--6 seconds in T4. We manually curate evaluation segments with the desired qualities -- high motion realism (T1), rhythmically aligned gestures (T2), attentive listening and dyadic adaptations (T3), and semantically meaningful expression (T4). This curation is an often-overlooked but essential step of evaluation design for two reasons: it ensures that the input segments are informative enough for the models to create high-quality aligned outputs, and it also raises the performance ceiling set by the motion-capture data.

\subsubsection{Monadic segments}
For T1 and T2, we use shared segments. Rather than enforcing a fixed segment duration, we include complete sentences of variable length. The goal of the segment selection process is to identify clips in which one speaker talks for most of the segment while the other remains largely silent. In addition, the main speaker should be actively gesticulating. Ideally, the selected samples should also come from a diverse set of interactions and speakers. Therefore, for these tasks, we extract only one segment per interaction.

For each interaction, we first load the Voice Activity Detection (VAD) annotations for both speakers in the conversation. We then merge VAD segments that are separated by less than 0.5 seconds to reduce recording artefacts and better capture complete utterances or sentences. To determine whether the main speaker is moving actively during a segment, we compute frame-to-frame differences in the Euler angles of the upper-body joints%
. Specifically, we calculate the absolute difference between consecutive frames and sum the motion values across all selected joints. The resulting value serves as an indicator of how much the person moved during the segment. %

Once the VAD has been processed and the motion information has been computed, we apply a sliding window over the entire interaction to identify segments in which one speaker talks for most of the window while the other remains silent, and where the main speaker also exhibits a substantial amount of movement that is likely to correspond to gesticulation. For each window, we compute the speaking time of both speakers to determine their roles (main speaker or listener). Since the listener is expected to remain mostly silent, we require that they speak for at most 10\% of the window duration (allowing for occasional non-relevant backchanneling), while the main speaker must speak for at least 50\% of the window duration to ensure the presence of continuous speech despite possible pauses.

After the main speaker has been identified, we use the transcript to align the segment boundaries with sentence boundaries. We detect the beginning and end of the sentences that overlap with the selected window. If a sentence exceeds the pre-set duration threshold of 20 seconds, it is discarded. After adjusting the boundaries, we compute the motion quantity for the frames contained in the segment. This sliding-window procedure is repeated across the entire interaction. Finally, among all candidate segments, we select the one with the highest motion value.

The selected segments are manually reviewed afterwards, since some may still contain artefacts such as trembling joints, twisted wrists, body penetrations, or audio issues. These problematic samples are discarded from the final selection. From the remaining set, we selected 170 samples from 50 different speakers for T1 and T2. Of these, 90 belong to the public subset and 80 to the private subset.

\subsubsection{Mismatched speech for monadic segments}
T2 requires mismatched segments, in which the motion remains unchanged while the audio is replaced with speech from a different part of the same interaction. To generate these mismatched samples, we process the interaction transcript again using a sliding window. As before, we aim to align the segment boundaries with complete sentences so that both the matched and mismatched segments begin at sentence boundaries. We then compare the VAD pattern of each candidate window with that of the matched segment and retain the candidate with the most similar VAD structure, ensuring that the mismatched audio segment does not overlap with the matched audio segment.

\subsubsection{Dyadic segments}
For T3, which focuses on listening behaviour, we follow a setup similar to that used for the previous tasks. The goal is still to obtain samples from a diverse set of interactions and speakers. However, due to the stricter selection requirements and the limited number of valid samples, we allow up to three non-overlapping segments per interaction.

As before, we process all interactions sequentially and preprocess the VAD annotations by merging segments that are separated by at most 0.5 seconds. We then compute the overlapping VAD regions between both speakers. Using a sliding-window approach, we calculate the percentage of time within each window during which both speakers are active simultaneously. We retain only those windows in which the overlap is between 30\% and 50\% of the window duration. These thresholds were determined empirically. We observed that, above the upper threshold, the segments often corresponded to situations in which the speakers were not engaging in a natural conversation, either because they were speaking over each other continuously or because of audio-related artefacts. The lower threshold was introduced to ensure that the listener still contributed actively to the interaction through backchanneling or verbal feedback directed at the main speaker. While active listening can also occur without speech, we did not identify suitable examples of this behaviour in the dataset, since most silent segments were associated with relatively static listeners.

For the windows that satisfy the VAD overlap requirement, we assign the listener role to the speaker with the lower speaking time. We then compute the amount of movement as described for T1 and T2, but considering only the listener's motion. For each interaction, we retain and render the three windows with the highest listener motion values. The rendered segments are manually reviewed, and only those in which the gestures of both speakers (especially those of the listener) are meaningful and relevant to the conversation are selected. In total, we selected 46 segments for evaluation from 41 different speakers. Of these, 24 belong to the public subset and 22 to the private subset.

\subsubsection{Mismatched interlocutor for dyadic segments}
To generate the mismatched pairs for the dyadic mismatching study, for each segment A, we first identify another segment B within the same subset (public or private) in which the agent has the most similar VAD to the agent of segment A. For the motion-capture condition, mismatched segments are created by keeping the interlocutor's audio and motion and the agent's audio from segment A unchanged, while replacing the agent's motion with the motion from segment B.

In the case of challenge submissions, challenge participants generate the agent's motion conditioned on the agent's audio, and the interlocutor's audio and motion from segment B. As a result, the generated motion is expected to correspond to a different conversation. To create the mismatched video for the user study, we combine the interlocutor's audio and motion from segment A with the agent's audio from segment A and the generated agent motion.

\subsubsection{Semantic segments}
Finally, for T4, we use the \emph{grounded gestures} subset, which contains recordings of a single speaker uttering individual sentences that include a word intended to be emphasised. These target words are marked with asterisks in the \texttt{interactions.csv} file, making them easy to identify.

Since each recording contains multiple unrelated sentences, the first step is to identify the beginning and end of every sentence containing a marked word. The selected segments are then rendered and manually inspected to discard samples containing motion or audio artefacts. After this filtering process, T4 includes 257 segments from 50 different speakers, of which 178 belong to the public subset and 79 to the private subset.

\subsection{User-study setup}
\subsubsection{Evaluation questions and response options}
We run four user studies, corresponding to the tasks in \cref{sec:methodology}. All four evaluations (shortened as E1--E4 below) use pairwise comparisons, posing the following questions to test-takers:
\begin{itemize}
    \item \textbf{E1 (realism):} \emph{In which video does the character gesture more like a real person?}
    \item \textbf{E2 (speech mismatching):} \emph{In which video do the character's movements fit the speech better?}
    \item \textbf{E3 (dyadic mismatching):} \emph{In which of the two videos does the character in the red T-shirt pay more attention to, and respond
              more appropriately to, the other person?}
    \item \textbf{E4 (semantic mismatching):} \emph{Which sentence is expressed by the character's gestures?}
\end{itemize}

In E1, E2, and E3, the response options are on a 5-point Likert-type rating scale, as in the GENEA Challenge 2023 \cite{kucherenko2023challenge} and the GENEA Leaderboard \cite{nagy2026towards}. In E4, there are four voting options, indicating which sentence (if any) the gestures express.

\subsubsection{Follow-up voting options (JUICE)}
Following the GENEA Leaderboard \cite{nagy2026towards}, we employ the \emph{JUICE} (``JUstify their choICE'') methodology \cite{girdhar2024factorizing} to collect more detailed feedback from test-takers. Whenever a test-taker indicates a preference (i.e., selects any option other than \emph{They Are Equal}), they are additionally asked to indicate \emph{which
    factors} drove that choice, by checking one or more attributes. The attribute lists are task-specific:
\begin{itemize}
    \item \textbf{E1 (realism):}
          \begin{itemize}
              \item Limbs or body penetrating each other, or physically impossible motion
              \item Glitches or artefacts (e.g.\ glitching wrists and fingers)
              \item The smoothness of the motion
              \item The amount and intensity of motion
              \item Recognisable gestures
          \end{itemize}
    \item \textbf{E2 (speech mismatching):}
          \begin{itemize}
              \item Fit the rhythm and timing of the speech better
              \item Emphasised the correct part (or parts) of the speech
              \item Better matched the content and meaning of the speech
              \item Better fit for the emotion of the speech
          \end{itemize}
    \item \textbf{E3 (dyadic mismatching):}
          \begin{itemize}
              \item Better turn-taking (motion respects who is speaking vs.\ listening)
              \item Reacts to the other person at the right moments
              \item More natural listening cues (nodding, small ``I'm following you'' movements)
              \item Coordinated movements with the other person (e.g.\ mirroring motion and poses, mimicry)
          \end{itemize}
\end{itemize}
An additional \emph{Other (please specify)} field lets test-takers describe factors not listed above. Note that we do not use JUICE for E4 due to the short length of the semantic segments.

\subsubsection{Crowdsourcing details}
We recruit test-takers through the \href{https://www.prolific.com}{Prolific} crowdsourcing platform. To incentivise high-quality ratings, we set the compensation level based on the median study duration, so that it corresponds to an hourly rate of approximately \pounds13.45, as quoted by the Living Wage Foundation in the UK. We also employ four attention checks per study to ensure that test-takers watch the videos and listen to the audio (if available) in full, and that they read the response options (in the case of the semantic study, where response options vary). All test-takers with failed attention checks are excluded from the results. \Cref{tab:study-sizes} summarises the number of stimuli, test-takers, and votes collected in each study.

\begin{table}[t]
    \caption{Number of stimuli (evaluation segments), test-takers, and votes collected in each of our four user studies.}
    \label{tab:study-sizes}
    \centering
    \small
    \setlength{\tabcolsep}{5pt}
    \renewcommand{\arraystretch}{1.15}
    \begin{tabular}{@{}lrrr@{}}
        \toprule
        Evaluation               & Stimuli & Test-takers & Votes   \\
        \midrule
        E1: Realism              & 170     & 154         & 3{,}789 \\
        E2: Speech mismatching   & 170     & 350         & 8{,}488 \\
        E3: Dyadic mismatching   & 46      & 159         & 3{,}895 \\
        E4: Semantic mismatching & 257     & 177         & 7{,}038 \\
        \bottomrule
    \end{tabular}
\end{table}

\section{Results}
\label{sec:results}

\subsection{Motion realism}
\label{sec:motion_realism_results}
\begin{figure}
    \centering
    \includegraphics[width=\linewidth]{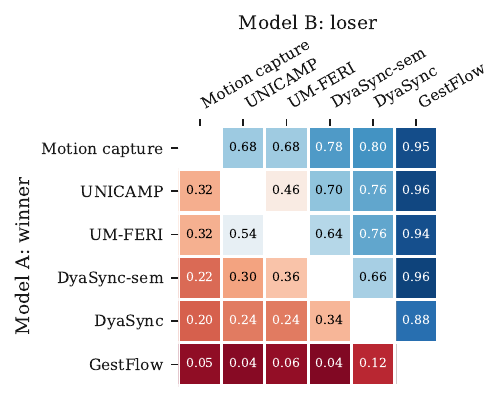}
    \caption{Motion-realism study: pairwise winrates, ignoring ties and the slight/clear preference distinction. Conditions are ordered by estimated Elo rating in descending order.}
    \label{fig:motion_realism_winfrac}
\end{figure}

The distribution of user-study responses is shown in \cref{fig:motion_realism_breakdown}, and the pairwise winrates between conditions are visualised in \cref{fig:motion_realism_winfrac}. Following \cite{nagy2026towards}, we compute Bradley-Terry Elo ratings from the responses by splitting ties and counting clear preferences twice; in \cref{fig:motion_realism_elo}, we report 95\% confidence intervals for the median Elo rating under non-parametric bootstrapping. Finally, for each challenge submission, the distribution of JUICE votes when compared against the motion-capture condition is reported in \cref{fig:motion_realism_juice}.

\subsection{Alignment with the speech}
\label{sec:speech_mismatch_results}
The distribution of responses for the speech mismatching study is shown in \cref{fig:speech_mismatch_breakdown}. We compute the \emph{speech appropriateness score} using the same bootstrapping methodology as before, and report the 95\% confidence intervals for the median score in \cref{fig:speech_mismatch_ratio}. Finally, the distribution of JUICE responses for each condition is shown in \cref{fig:speech_mismatch_juice}.

\subsubsection{Updated appropriateness scaling}
In previous GENEA Challenges \cite{kucherenko2023challenge}, the speech appropriateness score was defined as follows:
\begin{equation*}
    \mathrm{Appr}_{\text{speech}}^{\text{old}}
    = \frac{2\,C + S + \tfrac{1}{2}\,T}
    {2\,C + S + T + \overline{S} + 2\,\overline{C}},
\end{equation*}
where $C$, $S$, $T$, $\overline{S}$, and $\overline{C}$ denote the number of votes for \emph{clearly prefer matched}, \emph{slightly prefer matched}, \emph{they are equal}, \emph{slightly prefer mismatched}, and \emph{clearly prefer mismatched}, respectively. As in the realism study, clear preferences count twice and ties are split evenly, so the score from this formula lies in $[0, 1]$, with $0.5$ corresponding to chance performance.

However, in this year's challenge we introduce a semantic mismatching study, where it is more intuitive to set random performance at $0$, meaning that the model is not semantically expressive. Therefore, to enable consistent scaling between all appropriateness scores in the mismatching evaluation, we modify the scaling to be in the $[-1, 1]$ range, with chance performance at $0$:

\begin{equation*}
    \mathrm{Appr}_{\text{speech}}^{\text{updated}}
    = \frac{2\,C + S - \overline{S} - 2\,\overline{C} }
    {2\,C + S + T + \overline{S} + 2\,\overline{C}}.
\end{equation*}

This is a simple scaling change; the relationship between the two formulas is $\mathrm{Appr}_{\text{speech}}^{\text{updated}} = 2\,(\mathrm{Appr}_{\text{speech}}^{\text{old}} - 0.5)$.

\subsection{Alignment with the interlocutor}
The distribution of user responses collected in the dyadic mismatching study is shown in \cref{fig:dyadic_mismatch_breakdown}. Using the same methodology as in \cref{sec:speech_mismatch_results}, we report bootstrapped dyadic appropriateness scores in \cref{fig:dyadic_mismatch_ratio}, and the distribution of JUICE responses in \cref{fig:dyadic_mismatch_juice}.

\subsection{Semantic alignment}
\label{sec:semantic-results}
In \cref{fig:semantic_mismatch_breakdown}, we report the distribution of user-study responses in the semantic mismatching study. In comparison to the other two mismatching studies, which use a 5-point Likert-type rating scale, the semantic study does not distinguish between clear and slight preference, and it replaces the single tie option with two separate responses: \emph{neither is expressed} and \emph{both are expressed}.

Accordingly, the semantic appropriateness score is defined slightly differently to account for these differences:

\begin{equation*}
    \mathrm{Appr}_{\text{semantic}}
    = \frac{P - \overline{P}}
    {P + \overline{P} + N + B},
\end{equation*}

where $P$, $N$, $\overline{P}$, and $B$ denote the number of votes for \emph{matched text is expressed}, \emph{neither is expressed}, \emph{mismatched text is expressed}, and \emph{both are expressed}. Consistent with the other appropriateness scores, $\mathrm{Appr}_{\text{semantic}}$ lies in the $[-1, 1]$ range, with $0$ corresponding to chance performance.

The semantic appropriateness scores are reported in \cref{fig:semantic_mismatch_ratio}.

\section{Discussion}
\label{sec:discussion}

\begin{figure*}[h!]
    \centering
    \includegraphics[width=\linewidth]{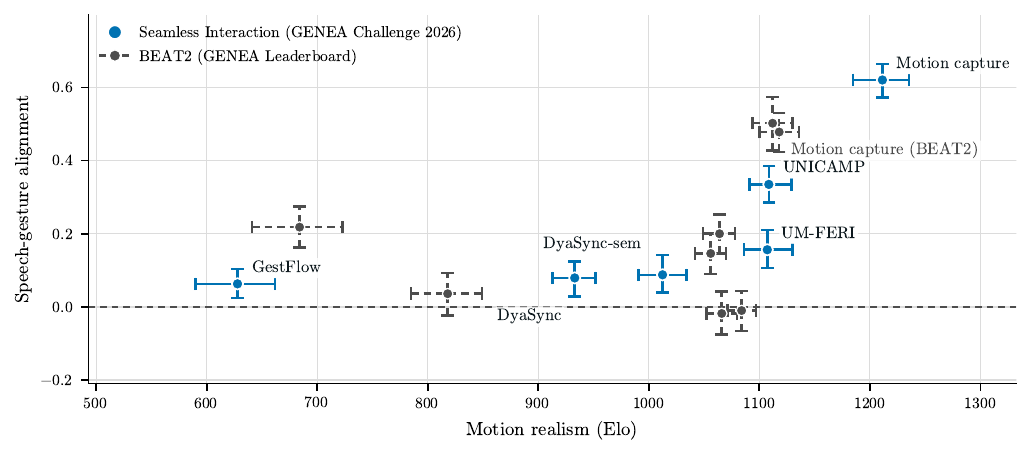}
    \caption{95\% confidence intervals for motion realism (Elo) and speech-gesture alignment (appropriateness score) for each condition in the two core evaluation criteria, comparing this challenge's results on Seamless Interaction with the GENEA Leaderboard results on BEAT2 (referenced in grey and without system names).}
    \label{fig:core-evaluation-vs-leaderboard}
\end{figure*}

Comparing our core evaluation results to those of the GENEA Leaderboard \cite{nagy2026towards} (\cref{fig:core-evaluation-vs-leaderboard}), we observe several notable differences. First, in terms of motion realism, challenge submissions are more separated from each other and from the motion-capture data than the systems evaluated on the BEAT2 leaderboard. While the submissions are different between the two evaluations, the scaling of the Elo scores is consistent, therefore these results are to some degree comparable. Second, we note that the Seamless motion-capture data achieves an Elo score about 100 higher than that of BEAT2, which could either imply that the filtered Seamless segments contain more expressive motion capture than the BEAT2 segments, or that the challenge submissions are overall weaker than the leaderboard systems.

Third, in terms of alignment with speech, the UNICAMP submission achieves a higher appropriateness score than any of the original six systems on the leaderboard, which demonstrates the utility of the dataset. This is further corroborated by the considerably higher appropriateness score of the Seamless motion-capture data compared to the prior result of the BEAT2 segments.

\begin{figure*}[h!]
    \centering
    \includegraphics[width=\linewidth]{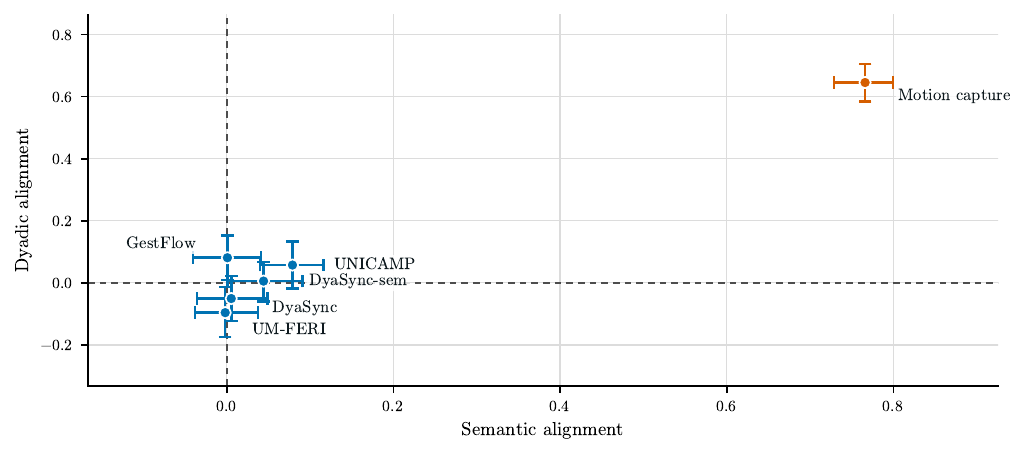}
    \caption{95\% confidence intervals for dyadic alignment versus semantic alignment (appropriateness scores) for each condition, comparing challenge submissions and the motion-capture reference across the two auxiliary evaluation criteria.}
    \label{fig:aux-evaluation-overview}
\end{figure*}

In terms of the auxiliary evaluation tasks, we find that none of the challenge submissions achieved convincing results (\cref{fig:aux-evaluation-overview}). However, the motion-capture segments were rated at a high level of alignment with the interlocutor, as well as for semantic alignment, which suggests that the dataset -- and the mismatching methodology coupled with careful selection of segments -- is suitable for evaluating these more advanced capabilities of gesture-generation systems moving forward.

We note that the mismatching evaluations provide a unique benefit in that they have a theoretical expected bottom-line performance of zero in the case of an input-independent system (i.e., one that does not use semantic or dyadic conditioning). This is in contrast with commonly used voting methodologies that compare two systems directly, which makes it difficult to ground the results in a meaningful way. Our dyadic evaluation results, perhaps counterintuitively, show that the $95\%$ confidence interval of the appropriateness score can fall below zero, even in the case of a monadic system like UM-FERI. There are many possible factors that may contribute to such a finding, including that the statistical analysis does not necessarily control for the effects of the random sample of system outputs (system-level randomness), nor for the effects of the particular selection of dyadic segments evaluated in the user study (study-level randomness). Both of these random factors are expected to have the greatest effect in the dyadic study, which uses the smallest number of speech segments, and their effect will not shrink towards zero even if the number of test-takers were to approach infinity. Given that many systems are evaluated -- some of which are cases where the null hypothesis (no impact of the interlocutor in the dyadic evaluation) is known to be true -- it is not statistically unlikely for one of the confidence intervals to wrongly exclude zero wholly or in part due to random chance. There is precedent for this in the GENEA Challenge 2023 \cite{kucherenko2023genea}.

\section{Conclusion}
\label{sec:conclusion}
In this preprint, we presented the 2026 GENEA Challenge, which focused on evaluating new gesture-generation systems on the Seamless Interaction dataset, both in terms of established criteria (motion realism and alignment with speech) and in terms of forward-looking capabilities (semantic and dyadic alignment). Our results indicate that the Seamless Interaction dataset may be promising as a benchmark for future gesture-generation models, and that the proposed evaluation methodology is suitable for assessing semantic and dyadic alignment in a disentangled manner. We hope that the challenge results will inspire future research on gesture generation, and that the collected votes and outputs, alongside the evaluation methodology, will be useful for the community.

\section*{Acknowledgements}
RN, SAG, MT, and GEH were partially supported by the Wallenberg AI, Autonomous Systems and Software Program (WASP) funded by the Knut and Alice Wallenberg Foundation. YY and GEH were partially supported by the Industrial Strategic Technology Development Program (grant no.\ 20023495) funded by MOTIE, Republic of Korea.

\clearpage
\raggedbottom
{
    \small
    \bibliographystyle{ieeenat_fullname}
    \bibliography{main}

\begin{thebibliography}{11}
\providecommand{\natexlab}[1]{#1}
\providecommand{\url}[1]{\texttt{#1}}
\expandafter\ifx\csname urlstyle\endcsname\relax
  \providecommand{\doi}[1]{doi: #1}\else
  \providecommand{\doi}{doi: \begingroup \urlstyle{rm}\Url}\fi

\bibitem[Agrawal et~al.(2025)Agrawal, Akinyemi, Alvero, Behrooz, Buffalini,
  Carlucci, Chen, Chen, Chen, Cheng, et~al.]{agrawal2025seamless}
Vasu Agrawal, Akinniyi Akinyemi, Kathryn Alvero, Morteza Behrooz, Julia
  Buffalini, Fabio~Maria Carlucci, Joy Chen, Junming Chen, Zhang Chen, Shiyang
  Cheng, et~al.
\newblock Seamless interaction: Dyadic audiovisual motion modeling and
  large-scale dataset.
\newblock \emph{arXiv preprint arXiv:2506.22554}, 2025.

\bibitem[Crnek(2026)]{crnek2026umferi}
Karlo Crnek.
\newblock {UM-FERI} approach to {GENEA} challenge 2026, 2026.
\newblock Non-archival paper, to appear in the Interactive Social Agents
  Workshop at ECCV 2026.

\bibitem[Fang et~al.(2026)Fang, Lu, Bao, and Liu]{fang2026dyasync}
Fengyi Fang, Ye Lu, Qian Bao, and Xudong Liu.
\newblock {DyaSync}: Disentangling speech identity and activity for dyadic
  co-speech gesture generation, 2026.
\newblock Non-archival paper, to appear in the Interactive Social Agents
  Workshop at ECCV 2026.

\bibitem[Girdhar et~al.(2024)Girdhar, Singh, Brown, Duval, Azadi, Rambhatla,
  Shah, Yin, Parikh, and Misra]{girdhar2024factorizing}
Rohit Girdhar, Mannat Singh, Andrew Brown, Quentin Duval, Samaneh Azadi,
  Sai~Saketh Rambhatla, Akbar Shah, Xi Yin, Devi Parikh, and Ishan Misra.
\newblock Factorizing text-to-video generation by explicit image conditioning.
\newblock pages 205--224, 2024.

\bibitem[Gomez~Sanchez and Costa(2026)]{gomez2026unified}
Johsac~Isbac Gomez~Sanchez and Paula D.~P. Costa.
\newblock A unified flow-matching {DiT} with direct continuous decoding for
  co-speech gesture generation.
\newblock In \emph{Proceedings of the European Conference on Computer Vision
  (ECCV) Workshops}, 2026.
\newblock To appear.

\bibitem[Guillen(2026)]{guillen2026gestflow}
Jos{\'e} Guillen.
\newblock {GestFlow}: A small flow-matching system for the {GENEA} challenge
  2026.
\newblock In \emph{Proceedings of the European Conference on Computer Vision
  (ECCV) Workshops}, 2026.
\newblock To appear.

\bibitem[Kucherenko et~al.(2021)Kucherenko, Jonell, Yoon, Wolfert, and
  Henter]{kucherenko2021large}
Taras Kucherenko, Patrik Jonell, Youngwoo Yoon, Pieter Wolfert, and Gustav~Eje
  Henter.
\newblock A large, crowdsourced evaluation of gesture generation systems on
  common data: The genea challenge 2020.
\newblock In \emph{26th international conference on intelligent user
  interfaces}, pages 11--21, 2021.

\bibitem[Kucherenko et~al.(2023{\natexlab{a}})Kucherenko, Nagy, Yoon, Woo,
  Nikolov, Tsakov, and Henter]{kucherenko2023challenge}
Taras Kucherenko, Rajmund Nagy, Youngwoo Yoon, Jieyeon Woo, Teodor Nikolov,
  Mihail Tsakov, and Gustav~Eje Henter.
\newblock The genea challenge 2023: A large-scale evaluation of gesture
  generation models in monadic and dyadic settings.
\newblock In \emph{Proceedings of the 25th International Conference on
  Multimodal Interaction}, page 792–801, New York, NY, USA,
  2023{\natexlab{a}}. Association for Computing Machinery.

\bibitem[Kucherenko et~al.(2023{\natexlab{b}})Kucherenko, Nagy, Yoon, Woo,
  Nikolov, Tsakov, and Henter]{kucherenko2023genea}
Taras Kucherenko, Rajmund Nagy, Youngwoo Yoon, Jieyeon Woo, Teodor Nikolov,
  Mihail Tsakov, and Gustav~Eje Henter.
\newblock The {GENEA} {C}hallenge 2023: A large-scale evaluation of gesture
  generation models in monadic and dyadic settings.
\newblock In \emph{Proceedings of the International Conference on Multimodal
  Interaction}, pages 792--801, 2023{\natexlab{b}}.

\bibitem[Kucherenko et~al.(2024)Kucherenko, Wolfert, Yoon, Viegas, Nikolov,
  Tsakov, and Henter]{kucherenko2024evaluating}
Taras Kucherenko, Pieter Wolfert, Youngwoo Yoon, Carla Viegas, Teodor Nikolov,
  Mihail Tsakov, and Gustav~Eje Henter.
\newblock Evaluating gesture generation in a large-scale open challenge: The
  {GENEA} {C}hallenge 2022.
\newblock \emph{ACM Transactions on Graphics (TOG)}, 2024.

\bibitem[Nagy et~al.(2026)Nagy, Voss, Hoang-Minh, Tsakov, Nikolov, Zhang, Ao,
  Yang, Huang, Cheng, et~al.]{nagy2026towards}
Rajmund Nagy, Hendric Voss, Thanh Hoang-Minh, Mihail Tsakov, Teodor Nikolov,
  Zeyi Zhang, Tenglong Ao, Sicheng Yang, Shaoli Huang, Yongkang Cheng, et~al.
\newblock Towards reliable human evaluations in gesture generation: Insights
  from a community-driven state-of-the-art benchmark.
\newblock In \emph{Proceedings of the IEEE/CVF Conference on Computer Vision
  and Pattern Recognition}, pages 2152--2164, 2026.

\end{thebibliography}
}

\clearpage
\appendix
\onecolumn
\section{Additional results}
\label{sec:appendix_figures}
This appendix collects the detailed per-study figures referenced in \cref{sec:results}: the breakdown of votes cast per condition, the bootstrapped ratings and appropriateness scores, and the JUICE justifications provided by test-takers.

\subsection{Motion realism}
\label{sec:appendix_motion_realism}

\begin{center}
    \includegraphics[width=\linewidth]{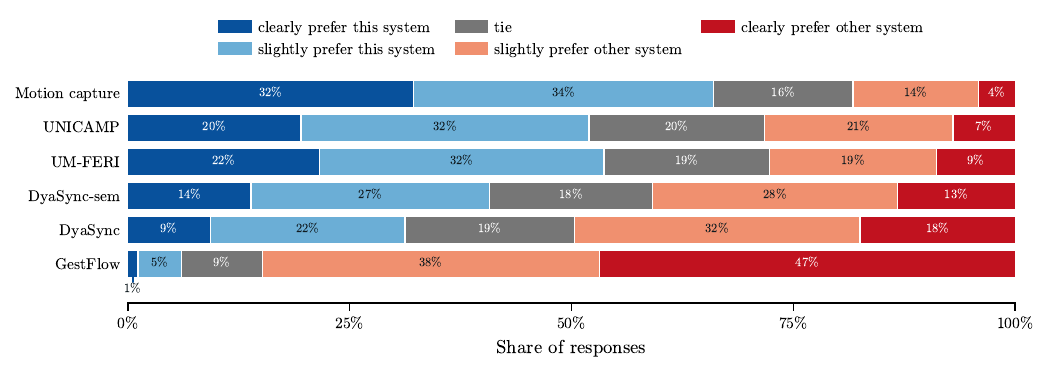}
    \captionof{figure}{Motion-realism study: breakdown of the votes cast per condition.}
    \label{fig:motion_realism_breakdown}
\end{center}

\begin{center}
    \includegraphics[width=\linewidth]{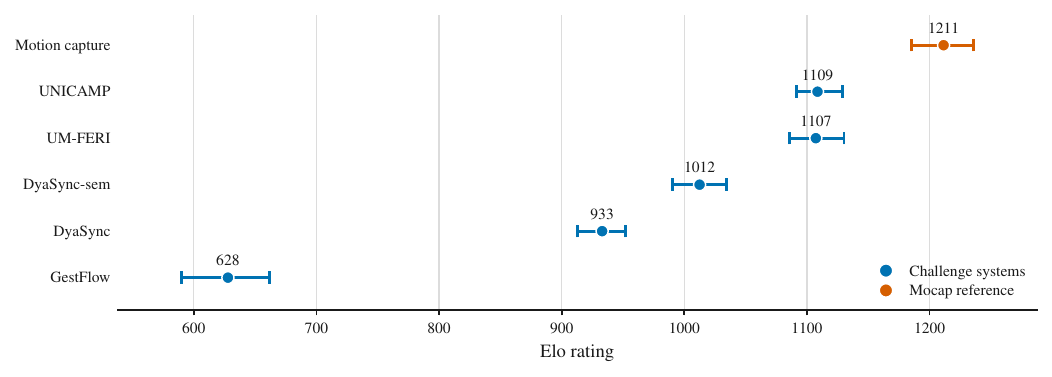}
    \captionof{figure}{Motion-realism study: Bradley-Terry Elo ratings for each condition, with 95\% confidence intervals obtained via bootstrapping. Conditions are ordered by estimated Elo rating in descending order.}
    \label{fig:motion_realism_elo}
\end{center}

\begin{center}
    \includegraphics[width=\linewidth]{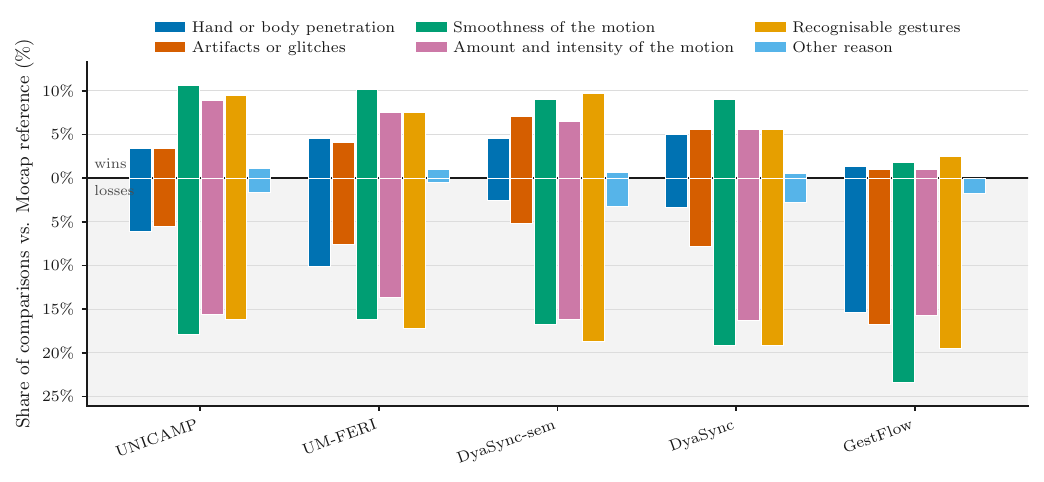}
    \captionof{figure}{Motion-realism study: JUICE justifications provided by test-takers.}
    \label{fig:motion_realism_juice}
\end{center}

\subsection{Alignment with the speech}
\label{sec:appendix_speech_mismatch}

\begin{center}
    \includegraphics[width=\linewidth]{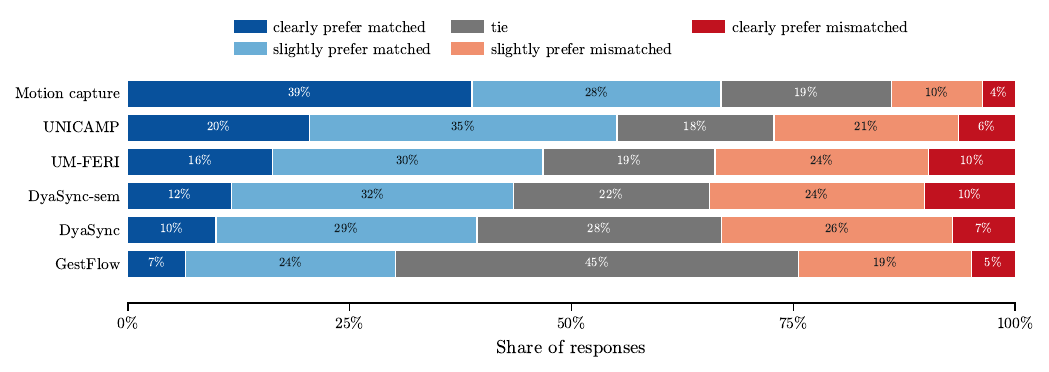}
    \captionof{figure}{Speech mismatching study: breakdown of the votes cast per condition.}
    \label{fig:speech_mismatch_breakdown}
\end{center}

\begin{center}
    \includegraphics[width=\linewidth]{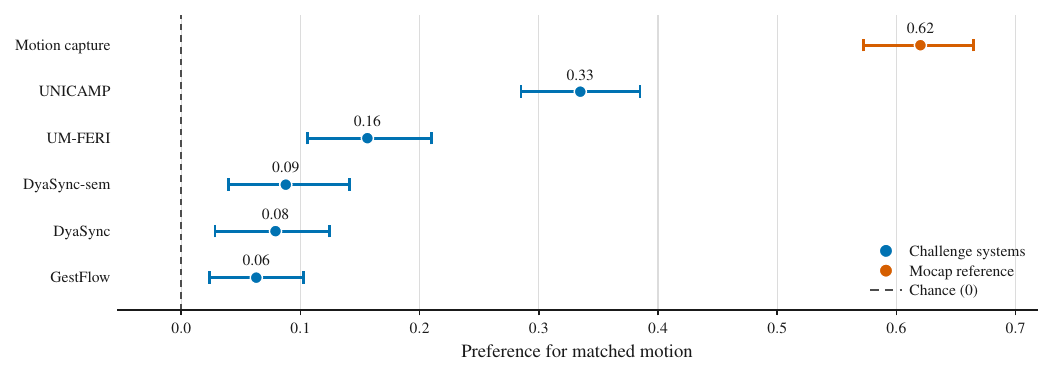}
    \captionof{figure}{Speech mismatching study: appropriateness score for each condition, with 95\% confidence intervals obtained via bootstrapping. Conditions are ordered by estimated appropriateness score in descending order.}
    \label{fig:speech_mismatch_ratio}
\end{center}

\begin{center}
    \includegraphics[width=\linewidth]{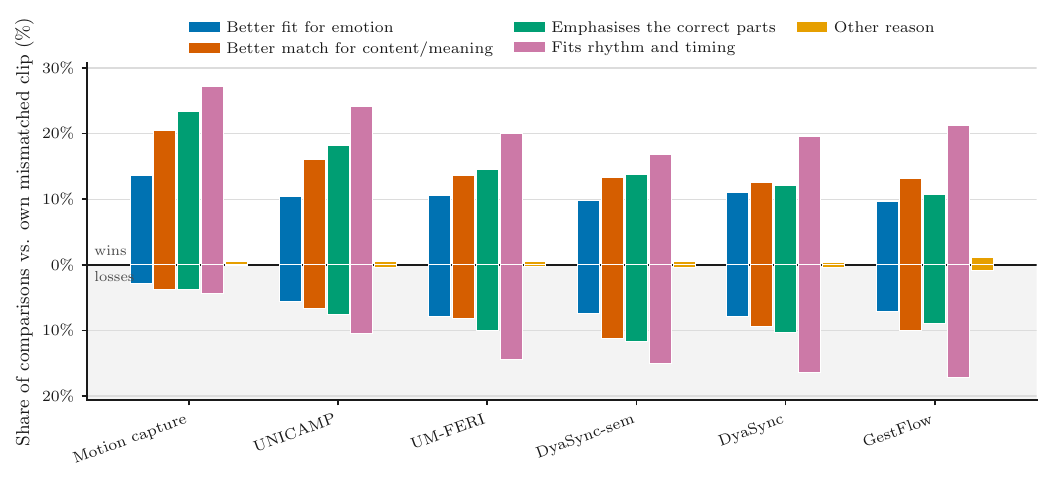}
    \captionof{figure}{Speech mismatching study: JUICE justifications provided by test-takers.}
    \label{fig:speech_mismatch_juice}
\end{center}

\subsection{Alignment with the interlocutor}
\label{sec:appendix_dyadic_mismatch}

\begin{center}
    \includegraphics[width=\linewidth]{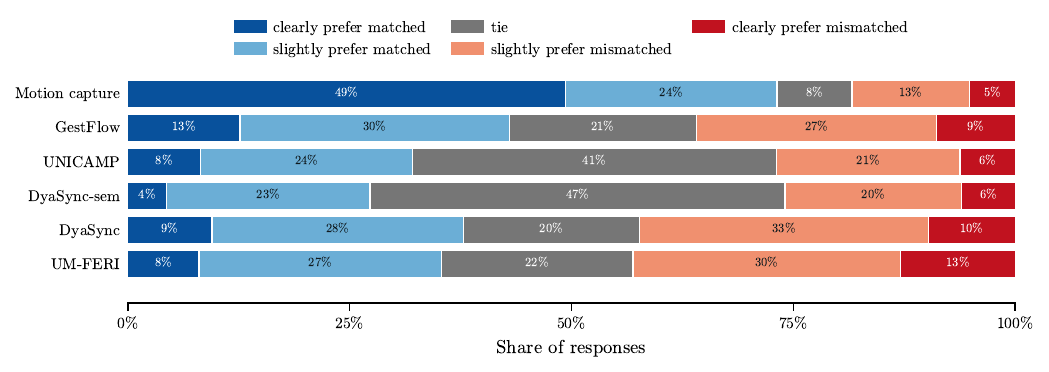}
    \captionof{figure}{Dyadic mismatching study: breakdown of the votes cast per condition.}
    \label{fig:dyadic_mismatch_breakdown}
\end{center}

\begin{center}
    \includegraphics[width=\linewidth]{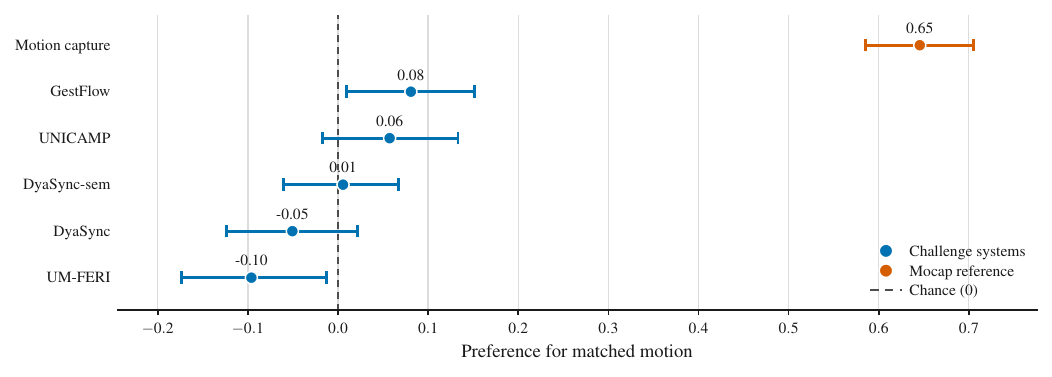}
    \captionof{figure}{Dyadic mismatching study: appropriateness score for each condition, with 95\% confidence intervals obtained via bootstrapping. Conditions are ordered by estimated appropriateness score in descending order.}
    \label{fig:dyadic_mismatch_ratio}
\end{center}

\begin{center}
    \includegraphics[width=\linewidth]{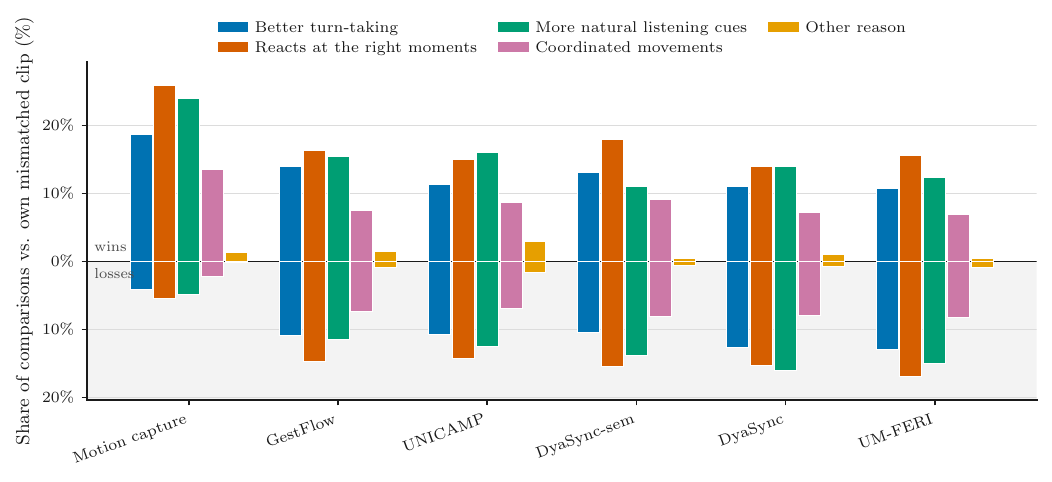}
    \captionof{figure}{Dyadic mismatching study: JUICE justifications provided by test-takers.}
    \label{fig:dyadic_mismatch_juice}
\end{center}

\subsection{Semantic alignment}
\label{sec:appendix_semantic_mismatch}

\begin{center}
    \includegraphics[width=\linewidth]{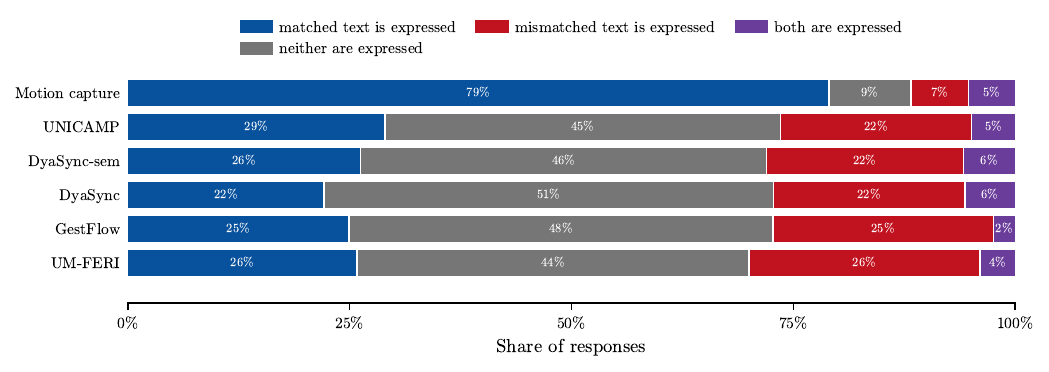}
    \captionof{figure}{Semantic mismatching study: breakdown of the votes cast per condition.}
    \label{fig:semantic_mismatch_breakdown}
\end{center}

\begin{center}
    \includegraphics[width=\linewidth]{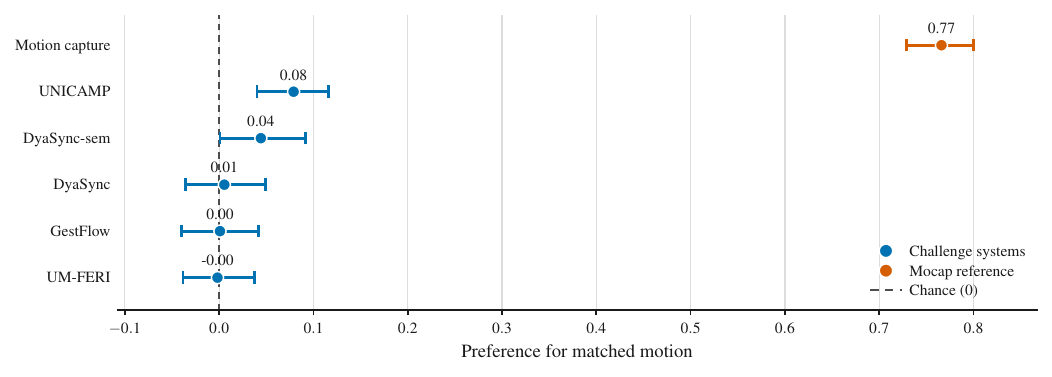}
    \captionof{figure}{Semantic mismatching study: appropriateness score for each condition, with 95\% confidence intervals obtained via bootstrapping. Conditions are ordered by estimated appropriateness score in descending order.}
    \label{fig:semantic_mismatch_ratio}
\end{center}

\end{document}